\pdfoutput=1

\documentclass[11pt]{article}

\usepackage[final]{acl}
\usepackage{acl}

\usepackage{times}
\usepackage{latexsym}
\usepackage{graphicx}
\usepackage{array}
\usepackage{pifont}
\usepackage[T1]{fontenc}

\usepackage[utf8]{inputenc}

\usepackage{microtype}

\usepackage{inconsolata}
\usepackage{float}
\usepackage{graphicx}
\usepackage{amsfonts,amssymb}
\usepackage{booktabs}
\usepackage{multirow}
\usepackage{makecell} 
\usepackage{bm}
\usepackage{amsmath}
\usepackage{tcolorbox}
\usepackage{xcolor}
\usepackage{setspace}
\usepackage{relsize}
\title{AgentsCourt: Building Judicial Decision-Making Agents with Court Debate Simulation and Legal Knowledge Augmentation}

\author{Zhitao He${}^{1,2}$, Pengfei Cao${}^{1,2}$, Chenhao Wang${}^{1,2}$, Zhuoran Jin${}^{1,2}$, Yubo Chen${}^{1,2}$\\
{\bf Jiexin Xu${}^{3}$ ,Huaijun Li${}^{3}$, Xiaojian Jiang${}^{3}$, Kang Liu${}^{1,2}$, Jun Zhao${}^{1,2}$} \\
         ${}^1$ The Laboratory of Cognition and Decision Intelligence for Complex Systems, 
         \\ Institute of Automation, Chinese Academy of Sciences, Beijing, China 
         \\${}^2$  School of Artificial Intelligence, University of Chinese Academy of Sciences, Beijing, China 
         \\${}^3$ AI Lab, China Merchant Bank, ShenZhen, China
         \\ \texttt{\{zhitao.he, pengfei.cao, yubo.chen, kliu, jzhao\}@nlpr.ia.ac.cn} }

\begin{document}
\maketitle
\begin{abstract}

With the development of deep learning, natural language processing technology has effectively improved the efficiency of various aspects of the traditional judicial industry. However, most current efforts focus on tasks within individual judicial stages, making it difficult to handle complex tasks that span multiple stages. As the autonomous agents powered by large language models are becoming increasingly smart and able to make complex decisions in real-world settings, offering new insights for judicial intelligence. In this paper, (1) we propose a novel multi-agent framework, \textit{AgentsCourt}, for judicial decision-making. Our framework follows the classic court trial process, consisting of court debate simulation, legal resources retrieval and decision-making refinement to simulate the decision-making of judge. (2) we introduce \textit{SimuCourt}, a judicial benchmark that encompasses 420 Chinese judgment documents, spanning the three most common types of judicial cases. Furthermore, to support this task, we construct a large-scale legal knowledge base, Legal-KB, with multi-resource legal knowledge. (3) Extensive experiments show that our framework outperforms the existing advanced methods in various aspects, especially in generating legal articles, where our model achieves significant improvements of 8.6\% and 9.1\% F1 score in the first and second instance settings, respectively.

\end{abstract}

\begin{figure}[t]
	 	\centering{ 
	 	\includegraphics[width=0.49\textwidth]{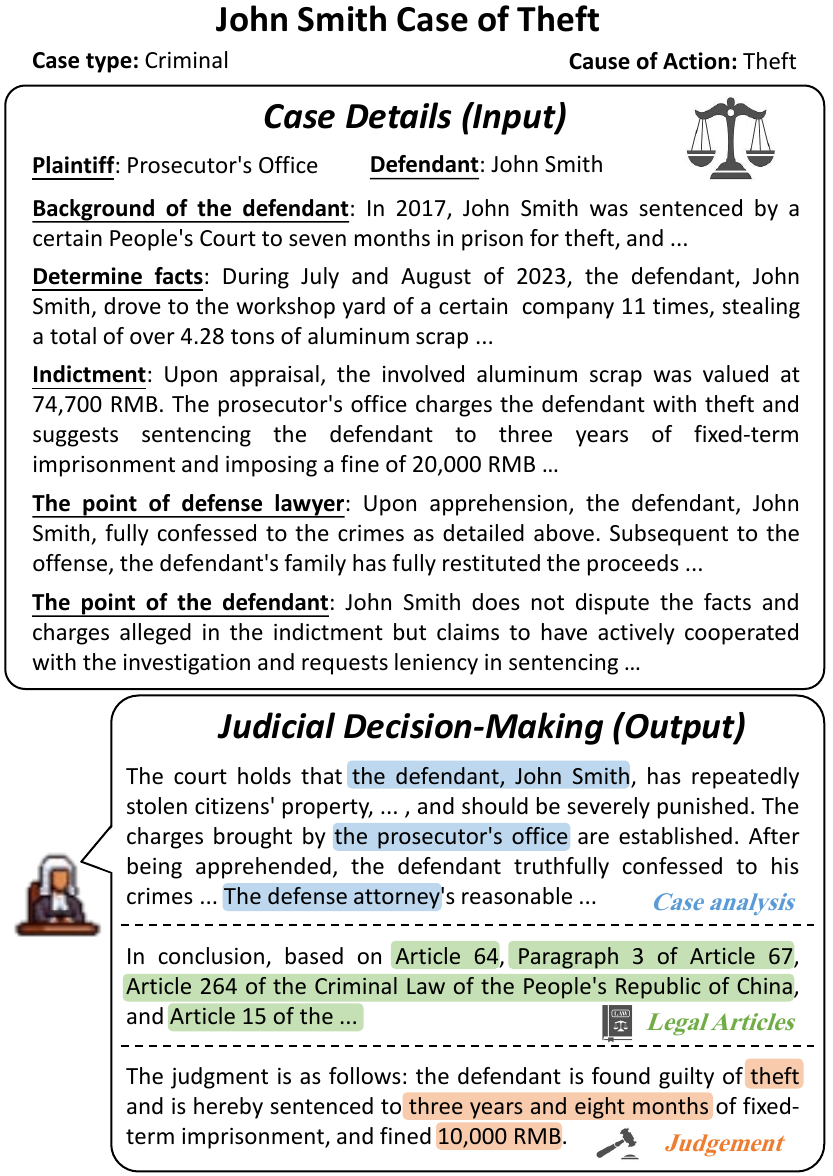}}  
	 	\caption{We formulate the Judicial Decision-Making task using the real-world judgement documents: given the case details above, judge agent must 1) conduct a logically clear case analysis; 2) provide precise legal articles; 3) issue a definitive judgement.}
	 	\label{fig:task_data}
\end{figure}

\section{Introduction}

\begin{table*}[t]
\centering
\resizebox{\linewidth}{!}{
\begin{tabular}{c|ccccc}
\toprule



\multicolumn{1}{l}{\multirow{2}{*}{\textbf{Framework}}} \vline & \textbf{AgentsCourt} & \textbf{LaWGPT} & \textbf{PLJP} & \textbf{HRN} & \textbf{RLJP}  \\

\multicolumn{1}{c}{} \vline & \textbf{(This work)} & \textbf{\citep{LaWGPT}}  & \textbf{\citep{wu2023precedent}} & \textbf{\citep{lyu2023multi}} & \textbf{\citep{wu2022towards}}\\

\midrule

 \multicolumn{1}{l}{Case Analysis}\vline & \textcolor{red}{\checkmark} & \textcolor{red}{\checkmark} & \textcolor[RGB]{80,80,80}{\ding{55}} & \textcolor[RGB]{80,80,80}{\ding{55}} & \textcolor{red}{\checkmark}  \\

\multicolumn{1}{l}{Precedent Retrieval}\vline & \textcolor{red}{\checkmark} & \textcolor[RGB]{80,80,80}{\ding{55}} & \textcolor{red}{\checkmark} & \textcolor[RGB]{80,80,80}{\ding{55}} & \textcolor[RGB]{80,80,80}{\ding{55}} \\

\multicolumn{1}{l}{Web Research}\vline & \textcolor{red}{\checkmark} & \textcolor[RGB]{80,80,80}{\ding{55}} & \textcolor[RGB]{80,80,80}{\ding{55}} & \textcolor[RGB]{80,80,80}{\ding{55}} & \textcolor[RGB]{80,80,80}{\ding{55}} \\

\multicolumn{1}{l}{Court Simulation}\vline & \textcolor{red}{\checkmark} & \textcolor[RGB]{80,80,80}{\ding{55}} & \textcolor[RGB]{80,80,80}{\ding{55}} & \textcolor[RGB]{80,80,80}{\ding{55}} & \textcolor[RGB]{80,80,80}{\ding{55}} \\

\multicolumn{1}{l}{Judgement Prediction}\vline & \textcolor{red}{\checkmark} & \textcolor{red}{\checkmark} & \textcolor{red}{\checkmark} & \textcolor{red}{\checkmark} & \textcolor{red}{\checkmark} \\

\multicolumn{1}{l}{Legal Articles Generation}\vline & \textbf{Multiple} & Single & Single & Single & Single \\

\multicolumn{1}{l}{Case Type}\vline & \textbf{Various} & Various & Crime & Crime & Crime  \\

\bottomrule
\end{tabular}
}
\caption{\label{tab:comparison} A comparison of our AgentsCourt to notable legal domain frameworks.}
\end{table*}

Recent advances in deep learning have significantly impacted the legal domain, with notable achievements in legal question answering \cite{zhong2020jec, khazaeli2021free,  cui2023chatlaw}, legal case retrieval \cite{sugathadasa2019legal, shao2020bert, li2023sailer, shao2023understanding} and legal judgment prediction \cite{xiao2018cail2018, chalkidis2019neural, wu2022towards, wu2023precedent}. These developments have effectively alleviated the long-standing issue in the judicial industry of "too many cases, too few legal professionals". However, case trial is a coherent process involving multiple stages such as court debates, case analysis, and  legal judgment prediction. The complexity of this process demands close collaboration and interaction between stages. Although current research has made progress in individual stages, it often overlooks the inherent connections between these stages of the trial process. This results in the need to rely on the deep involvement of legal experts when dealing with complex judicial decisions. Meanwhile, autonomous agents based on large language models (LLMs) have shown considerable progress in various traditional natural language processing (NLP) tasks \cite{brown2020language, wei2022chain, wang2023plan, qian2023communicative, wu2023autogen} and making decisions in real-world environments \cite{yao2023react, richards2023autogpt, chen2023agentverse}, which offers new insights for judicial intelligence.

However, simulating judicial decision-making is a non-trivial task because agents must navigate complex situations involving multiple stakeholders, understand the subtle nuances of legal provisions, and consider ethical and social justice factors. This presents three unique challenges to the agent system: (1) \textit{Intricate ethical relationships}. In judicial decision, ethical and moral considerations, which are often subtle and multi-faceted, must be taken into account.(2) \textit{Expert knowledge of judicial domain}. Judicial adjudication requires an in-depth understanding and accurate application of specialized knowledge such as laws, regulations and precedents. (3) \textit{Complex and hybrid reasoning}. The agents must be capable of handling a complex amalgamation of logical, factual, and legal reasoning, often interwoven in cases. 



\begin{figure}[t]
	 	\centering{ 
	 	\includegraphics[width=0.49\textwidth]{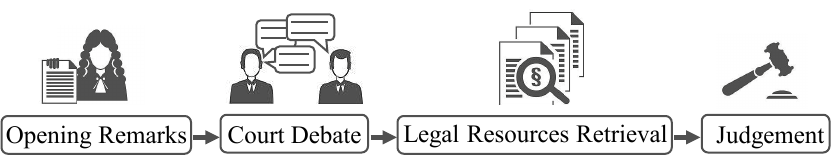}}  
	 	\caption{Simplified court trial process.}
	 	\label{fig:court_process}
\end{figure}

To tackle the aforementioned challenges, we propose a novel multi-agent framework, \textbf{AgentsCourt}, for the Judicial Decision-Making task. As illustrated in Figure \ref{fig:task_data}, given the case details, the task requires the agent to conduct a logically clear \textit{case analysis}, provide precise \textit{legal articles} and issue a definitive \textit{judgement}. AgentsCourt follows the classic court trial process: opening remarks, court debate, precedent retrieval, and judgement, as depicted in Figure \ref{fig:court_process}. Specifically, we first develop a \textit{Court Debate Simulation Module} with three agents, which serves as a platform for all parties involved to present their points to clarify the intricate ethical relationships in the case. One agent serves as the judge to open a court session and announce the basic facts of the case. The other two agents are designed as the plaintiff and the defendant respectively, and articulate their points of view during the court debate phase. Then, we devise the \textit{Legal Resources Retrieval Module} to address the inadequacy of expert knowledge. This module employs a judge assistant agent to integrate the most relevant precedents, articles and other information retrieved from the knowledge base we constructed and the internet. Next, we propose the \textit{Decision-Making Refinement Module} to facilitate complex and hybrid reasoning. This module firstly makes a preliminary judgement according to the inherent judicial expertise of the agent elicited by the established facts of current case and the transcripts of court debate, then subsequently refines the judgement using legal information retrieved. 

The comparison between our framework and prior works is listed in Table \ref{tab:comparison}. It is worth noting that our framework is not tailored to a specific legal system. AgentsCourt can achieve court simulation, precedent retrieval, judgment prediction, and supports the generation of multiple legal articles for practical judicial practice.

We also introduce \textbf{SimuCourt}, a judicial benchmark designed to evaluate Agent-as-Judge across a spectrum of different cases. SimuCourt encompasses 420 Chinese judgement documents, spanning the three most common types of judicial cases — criminal, civil, and administrative — in both first-instance and second-instance (appellate) courts, as well as covering three key societal roles: government agencies, the prosecutor's office, and individuals. Specifically, criminal cases involve acts that are identified as violations of criminal law, such as theft. Civil cases typically involve disputes between individuals, such as contract disputes or torts. Administrative cases concern disputes between individuals and government agencies. All the cases come from the China Judgements Online \footnote{\url{https://wenshu.court.gov.cn/}}, which is an official platform established by the Supreme People's Court of China, aimed at publicly releasing the judgement documents of courts at all levels in China. Furthermore, we construct a large-scale legal knowledge base, \textbf{Legal-KB}, to support this domain task. It encompasses a variety of legal knowledge, including effective laws and regulations, highly cited judicial papers, and precedents from recent years. The use of real data allows the agents developed on it can be transferred into real applications without any gaps. 





We summarize our contributions as follows:
\begin{itemize}
\item We propose a novel multi-agent framework AgentsCourt. Given the basic information of a case, our framework can sequentially simulate court debate, retrieve precedents, analyze cases, provide legal articles, and deliver clear judgment. The new judicial paradigm simplifies the process of making judicial decisions, significantly enhancing judicial efficiency.
\item We introduce SimuCourt, a judicial benchmark encompasses the three most common types of cases, enabling reliable assessment of the judicial analysis and decision-making power of agents for real judicial practice. Furthermore, we construct a legal knowledge base, Legal-KB, with multi-resource legal knowledge to support this task.

\item We perform extensive experiments and ablation studies. The results indicate that our framework outperforms the existing advanced methods in various aspects, especially in generating legal articles, where our system achieves notable improvements of 8.6\% and 9.1\% F1 score in the first and second instance experimental settings, respectively. 

\end{itemize}


\section{Related Work}






\noindent \textbf{Legal Artificial Intelligence} \quad 
Legal Artificial Intelligence seeks to improve legal tasks by employing artificial intelligence techniques \cite{surden2019artificial, zhong-etal-2020-nlp, katz2023natural}. With the continuous development of deep learning, the legal field has witnessed the emergence of more intelligent applications across various legal tasks. These tasks span across areas such as legal judgment prediction (LJP) \cite{xiao2018cail2018, zhong2018legal, xu2020distinguish, yue2021neurjudge, wu2022towards, wu2023precedent}, legal question answering \cite{zhong2020jec, cui2023chatlaw, louis2024interpretable, fei2023lawbench}, legal language understanding \cite{chalkidis-etal-2022-lexglue,xiao2021lawformer, niklaus2023lextreme, yu-etal-2023-exploring}, legal case retrieval \cite{sugathadasa2019legal, shao2020bert, li2023sailer, shao2023understanding}, legal document summarization \cite{kanapala2019text, jain2023bayesian, jain2024sentence}. While these existing efforts have made progress in individual legal tasks, they have overlooked the interconnection between different tasks, resulting in the necessity to heavily rely on the deep involvement of legal experts when dealing with complex judicial decisions. In this work, we focus on completing the entire process of judicial decision-making through multi-agent collaboration.

\noindent \textbf{Multi-agent framework} \quad Cooperation among agents like human group dynamics can enhance the efficiency and effectiveness of task accomplishment. \citet{li2023camel} enables two communicative agents to engage in a conversation and cooperate with each other to solve assigned tasks. \citet{park2023generative} found social behaviors autonomously emerge within a group of agents. \citet{qian2023communicative, hong2023metagpt} present innovative paradigms that leverages LLMs throughout the entire software development process by natural language communication. \citet{du2023improving, zhang2023building, he2023lego, chen2023agentverse, wu2023autogen} further leverage multi-agent cooperation to achieve better performance on multiple tasks. 


\subsection{Task Formulation}
We propose a generative task to evaluate agent as judge. Specifically, as shown in Figure \ref{fig:task_data}, we formulate the Judicial Decision-Making task as given the case details of a case, such as Determine facts, Complaint/Indictment, Statement of the plaintiff and the defendant, the agent system needs to make a complete judicial decision, which includes a clear and reasonable case analysis, rigorous legal articles, and definitive final judgement. SimuCourt encompasses two experimental settings:

\begin{figure*}[t]
	 	\centering{ 
	 	\includegraphics[width=1\textwidth]{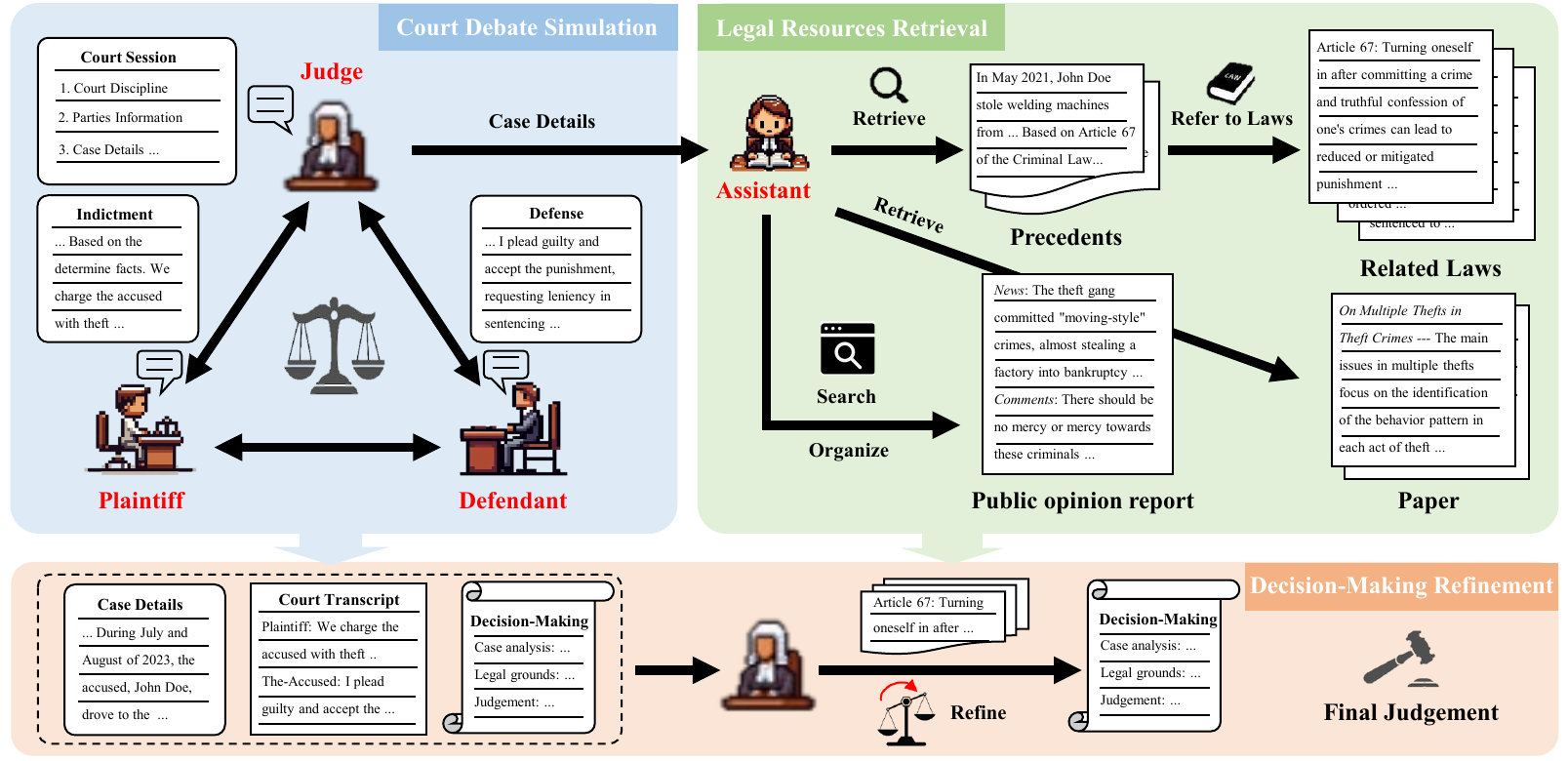}}  
	 	\caption{Overview of our multi-agent framework. The Court Debate Simulation Module recreates the court debate process through role-playing, mining different parties' points from limited real records. TheLegal Resources Retrieval Module employ an assistant agent to integrate information retrieved. The Decision-Making Refinement Module exploit the inherent judicial expertise of the judge agent and refines the judgment using information retrieved.
   }
	 	\label{fig:model}
\end{figure*}

\noindent \textbf{First Instance}\quad This setting refers to the trial court level, where the judge determines the guilt of the defendant, and assesses whether punitive measures are warranted. Within this setting, the primary focus is on evaluating the agent's understanding and analysis of case facts. 

\noindent \textbf{Second Instance}\quad This setting refers to the appellate court level. During this stage, the judge re-evaluates the case, considering new evidence. The objective at this stage is to ensure the legality and fairness of the initial judgement, identifying legal errors or inappropriate application of regulations from the first instance and demonstrating the capability to effectively handle new evidence. 

\section{The AgentsCourt Framework}

We propose a novel multi-agent framework, as shown in Figure \ref{fig:model}. Our framework is based on real-world court trial process and aims to study the collaboration of multiple agents, as well as how they contribute to judicial decision-making.


\subsection{Court Debate Simulation}
The court debate provides a platform for all parties involved to present their points and arguments comprehensively and fairly, which can significantly influence the judgement of the case.

\noindent \textbf{Court Simulation}\quad  Due to the majority of judgement documents only recording the key points of the plaintiff's and defendant's statements, obtaining complete court transcripts is challenging. Fortunately, as large language models have shown remarkable ability in role-playing \cite{li2023camel, qian2023communicative, chen2023agentverse}, in this module, we aim to reconstruct the court debate with multiple agents for each case.
We set up three agents to play the roles of the judge, plaintiff, and defendant respectively. For each agent, we carefully design an role-playing prompt to build their character personality and use the actual statements from judgment documents as the their starting prompts. It is worth noting that due to the limited record of statements in judgment documents, we combine the plaintiff and their representative, as well as the defendant and their representative, into the plaintiff and defendant, respectively, without setting separate roles for representatives. 

\noindent \textbf{Court Debate}\quad In this stage, both the plaintiff and the defendant need to present their arguments in line with their interests. The plaintiff should vigorously argue their complaint, articulating their stance and reasoning. Meanwhile, the defendant must defend their actions, aiming to prove their innocence or seek a lighter penalty. During the court session, the judge agent first delivers opening remarks, which include basic information about the plaintiff and the defendant, determination of facts, and so on. Then, the trial moves into the court debate stage and the communication between the agents will be recorded as court transcripts. We present an example of court transcripts in Table \ref{tab:transcript}.


\subsection{Legal Resources Retrieval}

Court debate serves as a platform to thoroughly explore the facts and contentious issues within a case, making the judge better comprehend the complexity of the matter. Furthermore, to make accurate judicial decisions, judges must possess extensive legal knowledge and case information. 

\noindent \textbf{Judge Assistant}\quad We assign an agent as judge assistant who is responsible for accessing the internet and the knowledge base. In terms of internet information acquisition, the assistant can use web research to seek open information, such as "Does the case have any public opinion?" This aids the judge in understanding the societal impact of the case and potential public perspectives. Ultimately, the agent organizes the retrieved news, comments to the judge, supporting the judge in making rational and well-founded judicial decisions.

\begin{figure}[t]
	 	\centering{ 
	 	\includegraphics[width=0.49\textwidth]{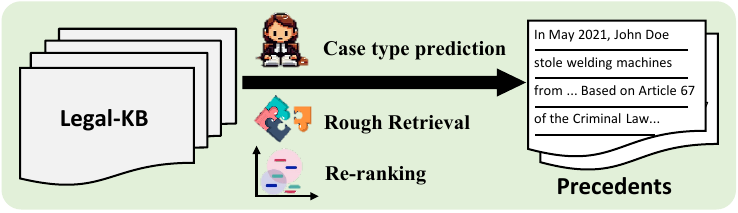}}  
	 	\caption{Automatic retrieval of precedents.}
	 	\label{fig:retrieve}
\end{figure}

\noindent \textbf{Automatic Information Retrieval}\quad In terms of knowledge base retrieval, as presented in Figure \ref{fig:retrieve}, the assistant first predict the type of case based on the determine facts of the current case. Due to the vast number of documents in the knowledge base, and the fact that cases with the same cause often have more similar keywords, we employs the BM25 model \cite{Lin_etal_SIGIR2021_Pyserini} for efficient \textit{rough retrieval} to obtain the top 100 documents from the knowledge base. Building on this, we further utilize the BGE-Large model \cite{bge_embedding} to encode and \textit{re-rank} these retrieved documents and choose the most relevant document to the current case as the optimal precedent. Additionally, to obtain more comprehensive laws and regulations relevant to the current case without introducing additional context, the judge assistant extracts the corresponding legal articles from the top 5 precedents as related legal provisions of current case. 


\subsection{Judgement Refinement}
In this module, we first exploit the inherent judicial expertise of the agent by utilizing determine facts of current case and transcripts of court debate to make a preliminary judgment. Then, the judge agent refines the judgment using information retrieved. 

\noindent \textbf{Preliminary Judgement}\quad As shown in the bottom of Figure \ref{fig:model}, after receiving the determine facts of current case and transcripts of simulated court debate, the judge agent takes the action of analysis, then provides its legal articles and subsequently reaching a preliminary judgement. 

\noindent \textbf{Judgement Refinement}\quad After obtaining the preliminary judgement which involves analyzing the specific details of the case, the judge agent uses precedent and relevant legal information from the assistant to refine the its judgement and provide the final judgement. This includes but is not limited to analyzing the precedent, referring to legal regulations and considering opinions of public. 

\section{The SimuCourt Benchmarck}

The task, Judicial Decision-Making, requires agents to conduct case analysis, generate legal articles and judgments. However, most existing legal datasets suffer from several limitations when it comes to assessing the Agent-as-Judge paradigm: 1) only contain the factual information of cases; 2) only focus on criminal cases; 3) only evaluate judgments. To this end, we propose SimuCourt, a judicial benchmark for a reliable assessment of the judicial analysis and decision-making power of agents. A comparison between our dataset and previous works is presented in Table \ref{tab:dataset_comp}.

\subsection{Data Collection}
We collect 420 real-world cases from the China Judgements Online, which span across two fundamental trial stages: first instance and second instance. These cases encompass three types: criminal, civil, and administrative. For first-instance cases, each sample includes the indictment, the point of the defendant, determine facts, etc. For second-instance cases, each sample contains petition for appeal, the point of the appellant and appellee, etc. Detailed list and data examples can be found in the Appendix \ref{sec:list}. Most of cases were released after April 2023. This minimizes the risk of data leakage\footnote{The cutoff date of pretraining data for gpt-3.5-turbo-0613 and gpt-4-1106-preview is officially before April 2023.}. 
Detailed data statistics of SimuCourt are shown in Table \ref{tab:case_type}. Furthermore, our dataset undergo rigorous scrutiny, ensuring the accuracy and completeness of the legal texts and information. Details of data collection and quality inspection can be found in Appendix \ref{data_appendix}.







\begin{table}
\centering
\resizebox{\linewidth}{!}{
\begin{tabular}{l|ccc}
\toprule

\textbf{Resource} & \textbf{SimuCourt} & \textbf{CAIL} & \textbf{SLJA-SYN}\\

\midrule

Background of Defendant? & \textcolor{red}{\checkmark} &\textcolor[RGB]{80,80,80}{\ding{55}} & \textcolor[RGB]{80,80,80}{\ding{55}} \\

Statement of Different Parties? & \textcolor{red}{\checkmark} & \textcolor[RGB]{80,80,80}{\ding{55}} & \textcolor[RGB]{80,80,80}{\ding{55}} \\

Multi-article Scenario? & \textcolor{red}{\checkmark} & \textcolor[RGB]{80,80,80}{\ding{55}} & \textcolor[RGB]{80,80,80}{\ding{55}}\\ 

Case Analysis Evaluation? & \textcolor{red}{\checkmark} & \textcolor[RGB]{80,80,80}{\ding{55}} & \textcolor{red}{\checkmark}\\ 

Judgement Evaluation & \textcolor{red}{\checkmark} & \textcolor{red}{\checkmark} & \textcolor{red}{\checkmark}\\ 

Laws Involved? & \textbf{443} & 1 & 1 \\

Case Retrival? & \textbf{6.5M} & 2.6M & \textcolor[RGB]{80,80,80}{\ding{55}} \\

Various Case Types? & \textbf{Crime, Civial, Admini.}& Crime & Crime \\

Different Instances Involved? & \textbf{First/Second} & First & First  \\ 

\bottomrule

\end{tabular}}
\caption{A comparison of our SimuCourt to remarkable legal domain datasets. CAIL \cite{xiao2018cail2018} is a widely used legal judgment prediction dataset, where each case comes with a fact description; SLJA-SYN \cite{deng2023syllogistic} is a comprehensive legal dataset designed to support multiple tasks such as article retrieval, article interpretation generation, criminal element generation and legal judgment prediction.}
\label{tab:dataset_comp}
\end{table}

\begin{table}
\centering
\resizebox{\linewidth}{!}{
\begin{tabular}{l|ccc}
\toprule

\textbf{Feature} & \textbf{Criminal} & \textbf{Civil} & \textbf{Administrative}\\

\midrule

\# of Cases & 140 &140&140 \\
\# of Causes of action & 44&51&33 \\
Avg \# of Legal articles & 6.3&3.3&1.6 \\ 
Max \# of Legal articles & 11&10&8 \\
Total \# of Legal articles & 198&153&92 \\ 
Avg. Length of Facts & 468.7&487.5&673.3 \\ 
Avg. Length of Analysis & 346.3&486.1&722.7 \\ 
Avg. Length of Cases & 2362.6&2473.8&3315.5 \\ 

\bottomrule

\end{tabular}}
\caption{Statistics of SimuCourt. Length is measured via the number of words}
\label{tab:case_type}
\end{table}

\subsection{Legal Knowledge Base Construction}


To make accurate judicial decisions, judges must possess extensive legal knowledge. Furthermore, given the diversity and complexity of human society, each case may involve different facts, parties, and locations. To this end, we construct a large scale legal knowledge base consists of laws, regulations, judicial interpretation, journal articles, and precedents.
Detailed data statistics of \textit{Legal-KB} are shown in Table \ref{tab:KB}.

\noindent \textbf{Laws, Regulations and Judicial interpretations}\quad We download various legal documents from the National Laws and Regulations Database of China\footnote{\url{https://flk.npc.gov.cn}}, an authoritative resource for legal information that includes national laws, administrative regulations, local regulations, and judicial interpretations. We remove legal documents that are no longer in effect. 


\noindent \textbf{Journal Articles}\quad Journal articles, typically authored by legal experts, can provide in-depth analysis and unique perspectives on specific legal issues. We collect highly-cited journal articles from 2010 to 2023 from the Chinese Legal Resources Knowledge Database \footnote{\url{https://lawnew.cnki.net/}}. 


\noindent \textbf{Precedents}\quad We collect all judgement documents of criminal, civil and administrative cases from the China Judgements Online for the years 2017 to 2022. However, as illustrated in Figure \ref{data_analysis} in the Appendix, the data exhibits a significant long-tail distribution. To balance the type of case, we limit the number of cases for each cause of action to no more than 20k. For those causes of action with more cases, we retain only the top 20k cases with the longest text as representatives of complex cases. 


\begin{table}[t]
\centering
\resizebox{\linewidth}{!}{
\begin{tabular}{lccc}
\toprule
\textbf{Type} & \textbf{Num} & \textbf{Tokens} & \textbf{Avg. Tokens}\\
\midrule
Laws and Regulations & 9K & 66M & 7390\\
Journal Articles & 29K & 15M & 521\\
Precedents & 6.5M & 27.1B & 4111 \\ 
\bottomrule
\end{tabular}}

\caption{Statistics of our legal knowledge base.}
\label{tab:KB}
\end{table}

\section{Experiments}

\begin{table*}
\centering
\resizebox{\linewidth}{!}{
\begin{tabular}{cc|ccc|ccc|ccc|ccc}
\toprule
\multicolumn{1}{c}{\multirow{3}{*}{}} &
\multicolumn{1}{l}{\multirow{3}{*}{\textbf{Model}}} \vline & 
\multicolumn{3}{c}{\multirow{2}{*}{\textbf{Legal Articles}}} \vline & 
\multicolumn{6}{c}{\multirow{1}{*}{\textbf{Judgement Results}}} \vline & 
\multicolumn{3}{c}{\multirow{2}{*}{\textbf{Case Analysis}}} \\

\cmidrule(lr){6-11}
\multicolumn{1}{c}{} & \multicolumn{1}{c}{} \vline & \multicolumn{3}{c}{} \vline & \multicolumn{3}{c}{Civil and Admini.} \vline & \multicolumn{3}{c}{Criminal} \vline & \multicolumn{3}{c}{}\\

\cmidrule(lr){3-14}
\multicolumn{1}{c}{} & \multicolumn{1}{c}{} \vline & \textbf{P} & \textbf{R} & \textbf{F} & \textbf{P} & \textbf{R} & \textbf{F} & \textbf{Charge} & \textbf{Prison term} & \textbf{Fine} & \textbf{Correctness} & \textbf{Logicality} & \textbf{Concision}\\

\midrule

 \multicolumn{1}{c}{\multirow{6}{*}{\rotatebox{90}{First}}} & \multicolumn{1}{l}{\textbf{GPT-3.5}}\vline & 0.127 & 0.109 & 0.117 & 0.367 & 0.498 & 0.423 & 0.822 & 0.253 & 0.412 & 0.466&0.510&0.493\\
 \multicolumn{1}{c}{} &\multicolumn{1}{l}{\textbf{GPT-4}}\vline & 0.139 & 0.133 & 0.136 & 0.398 & 0.559 & 0.465 & 0.875 & 0.287 & 0.462 & 0.503&0.553&\textbf{0.543}\\
 \multicolumn{1}{c}{} &\multicolumn{1}{l}{\textbf{ReAct}}\vline & 0.161 & 0.109 & 0.131 & 0.387 & 0.532 & 0.448 & 0.866 & 0.262 & 0.437 & 0.516&0.567&0.533\\
 \multicolumn{1}{c}{} &\multicolumn{1}{l}{\textbf{AutoGPT}}\vline & 0.171 & 0.123 & 0.143 & 0.392 & 0.543 & 0.455 & 0.862 & 0.275 & 0.450 & 0.523&0.576&0.520\\
 
  \multicolumn{1}{c}{} &\multicolumn{1}{l}{\textbf{LaWGPT}}\vline & 0.183 & 0.105 & 0.133 & 0.414 & 0.548 & 0.471 & 0.875 & 0.237 & 0.425 & 0.506 & 0.546 & 0.533\\
  
 \multicolumn{1}{c}{} &\multicolumn{1}{l}{\textbf{AgentsCourt}}\vline & \textbf{0.219} & \textbf{0.189} & \textbf{0.203} & \textbf{0.437} & \textbf{0.603} & \textbf{0.507} & \textbf{0.887} & \textbf{0.337} & \textbf{0.500} & \textbf{0.550}&\textbf{0.596}&0.526\\

 \midrule
 \midrule

 \multicolumn{1}{c}{\multirow{6}{*}{\rotatebox{90}{Second}}} & \multicolumn{1}{l}{\textbf{GPT-3.5}}\vline & 0.206 & 0.169 & 0.186 & 0.317 & 0.429 & 0.365 & 0.716 & 0.166 & 0.516 & 0.496&0.540&0.526\\
 \multicolumn{1}{c}{} &\multicolumn{1}{l}{\textbf{GPT-4}}\vline & 0.200 & 0.267 & 0.228 & 0.356 & 0.482 & 0.409 & 0.800 & 0.183 & 0.533& 0.530&0.583&0.576\\
\multicolumn{1}{c}{} & \multicolumn{1}{l}{\textbf{ReAct}}\vline & 0.209 & 0.235 & 0.221 & 0.364 & 0.457 & 0.405 & 0.800 & 0.150 & 0.516 & 0.526&0.586&0.570\\
 \multicolumn{1}{c}{} &\multicolumn{1}{l}{\textbf{AutoGPT}}\vline & 0.217 & 0.248 & 0.231 & 0.371 & 0.478 & 0.417 & 0.816 & 0.166 & 0.550 & 0.540&0.590&0.583\\

  \multicolumn{1}{c}{} &\multicolumn{1}{l}{\textbf{LaWGPT}}\vline & 0.225 & 0.231 & 0.227 & 0.382 & 0.472 & 0.422 & 0.850 & 0.133 & 0.483 & 0.503 & 0.553 & 0.566\\
  
 \multicolumn{1}{c}{} &\multicolumn{1}{l}{\textbf{AgentsCourt}}\vline & \textbf{0.271} & \textbf{0.284} & \textbf{0.277} & \textbf{0.400} & \textbf{0.528} & \textbf{0.456} & \textbf{0.833} & \textbf{0.200} & \textbf{0.583} & \textbf{0.583}&\textbf{0.633}&\textbf{0.593}\\

\bottomrule
\end{tabular}
}
\caption{\label{tab:result} Overall performance of our framework and baselines in the first and second instance experimental settings.}
\end{table*}

\subsection{Automatic Evaluation}
As example data illustrated in Table \ref{tab:criminal}, the legal articles and judgement are concise and structured. Therefore, we propose corresponding metrics for legal articles and judgement evaluation.


\noindent \textbf{Legal Articles Evaluation}\quad The correct legal articles is crucial for a fair judgment. Thus, we employ the strict matching method to assess the legal articles generated by the agent system. Specifically, we calculate the number of entries that match and do not match between the legal articles list of the agent system and the reference legal articles list. These counts are then micro-averaged to determine the overall precision, recall and F1 scores. Details can be found in Table \ref{tab:comp_article}.

\noindent \textbf{Judgement Evaluation for Civil and Administrative Cases}\quad The judgment of each civil or administrative case may encompass multiple results. While each result typically revolves around a single key point, it may involve specific monetary amounts and interest rate information. Consequently, traditional text matching methods based on similarity struggle to accurately capture these key points. Thus, we employ GPT-4 as an evaluator. Specifically, we separately count the number of matching and non-matching key points in the agent system's judgment results compared to the reference judgment results. The micro-averaged counts are used to calculate the overall precision, recall and F1 scores. Details is presented in Table \ref{tab:GPT_eval}.

\noindent \textbf{Judgement Evaluation for Criminal Cases}\quad Different from other cases, the sentence of criminal case typically include three core elements: charge, prison term, and fine. The determination of the charge must match the facts of the case. The specific amounts of the prison term and fines are based not only on the facts but also take into account the defendant's performance in court, including their attitude towards the crime and the defense they present for their actions. We calculate the accuracy of the agent system separately for these three items.

\subsection{Human Evaluation} The case analysis entails intricate logical reasoning and ethical considerations that are challenging to evaluate through automatic metrics or GPT-4. For each setting, we present a panel of three graduate students majoring in law a random sample of 100 entries from each setting and the following binary True/False criteria guidelines:
1) \textbf{Correctness}: Mark true if and only if the analysis is satisfying and considers all parties involved. 2) \textbf{Logicality}: Mark false if the analysis contains any illogical or untrue reasoning. 3) \textbf{Concision}: Mark true if the analysis covers all necessary information without any extra information. 

\begin{table*}
\centering
\resizebox{0.88\linewidth}{!}{
\begin{tabular}{c|c|c|ccc}
\toprule
\multicolumn{1}{c}{\multirow{2}{*}{\textbf{Model}}} \vline & 
\multicolumn{1}{c}{\multirow{2}{*}{\textbf{Legal Articles}}} \vline & 
\multicolumn{4}{c}{\multirow{1}{*}{\textbf{Judgement Results}}} \\

\cmidrule(lr){3-6}
\multicolumn{1}{c}{} \vline  & \multicolumn{1}{c}{} \vline & \multicolumn{1}{c}{Civil and Admini.} \vline & \multicolumn{1}{c}{Charge}  & \multicolumn{1}{c}{Prison term}  & \multicolumn{1}{c}{Fine}\\

\midrule

\multicolumn{1}{l}{\textbf{SimuCourt}}\vline & 0.203 & 0.507 & 0.887 & 0.337 & 0.500 \\
\multicolumn{1}{l}{\textbf{\quad w/o Court simulation}}\vline & 0.171 & 0.473 & 0.875 & 0.300 & 0.462 \\
\multicolumn{1}{l}{\textbf{\quad w/o Knowledge base}}\vline & 0.145 & 0.462 & 0.850 & 0.312 & 0.475 \\
\multicolumn{1}{l}{\textbf{\quad w/o Web search}}\vline & 0.196 & 0.488 & 0.865 & 0.325 & 0.487 \\

\bottomrule
\end{tabular}
}
\caption{\label{tab:abla} Ablation study of our framework in the first instance setting.}
\end{table*}

\subsection{Baselines}

\noindent \textbf{Vanilla} \quad We employ \texttt{gpt-3.5-turbo-1106} and \texttt{gpt-4-1106-preview} with few-shot as vanilla models. Furthermore, due to limited budget, we only use the \texttt{gpt-3.5-turbo-1106} as foundation models of all agent systems.

\noindent \textbf{ReAct} \cite{yao2023react} \quad This system enables the agent to improve its actions based on the outcomes of past activities like searches or tool usage.

\noindent \textbf{AutoGPT} \cite{richards2023autogpt} \quad This is the most advanced agents framework, incorporating a variety of tools and prompts designed to facilitate the automatic planning and execution of specified tasks.

\noindent \textbf{LaWGPT} \cite{LaWGPT} \quad This is currently the most popular Chinese legal large language model\footnote{\url{https://github.com/pengxiao-song/LaWGPT}}, which has undergone extensive pre-training on Chinese legal corpora and fine-tuning on legal instructions, based on the general Chinese foundation model (Chinese-LLaMA-7B). It possesses strong capabilities in understanding and generating legal content.

\subsection{Main Results}


As shown in Table \ref{tab:result}, our framework outperforms other models in all aspects. For the evaluation on legal articles, our proposed framework achieved performance improvements of 8.6\% and 9.1\% in the two experimental settings, respectively. In contrast, GPT-4's performance in the first and second instance settings only reach 13.6\% and 22.8\%, respectively. This not only indicates significant shortcomings in the capabilities of LLMs in sourcing legal provisions, but also reflects the high challenge of our benchmark. In terms of judgment results evaluation, while all models performed well in the conviction of criminal cases, there is still a significant gap in determining prison term and fines compared to standard results. Furthermore, although the analysis of these systems has shown a certain degree of logicality, there is still room for improvement in terms of correctness and concision.

\subsection{Discussion and Analysis}

\begin{figure}[t]
	 	\centering{ 
	 	\includegraphics[width=0.49\textwidth]{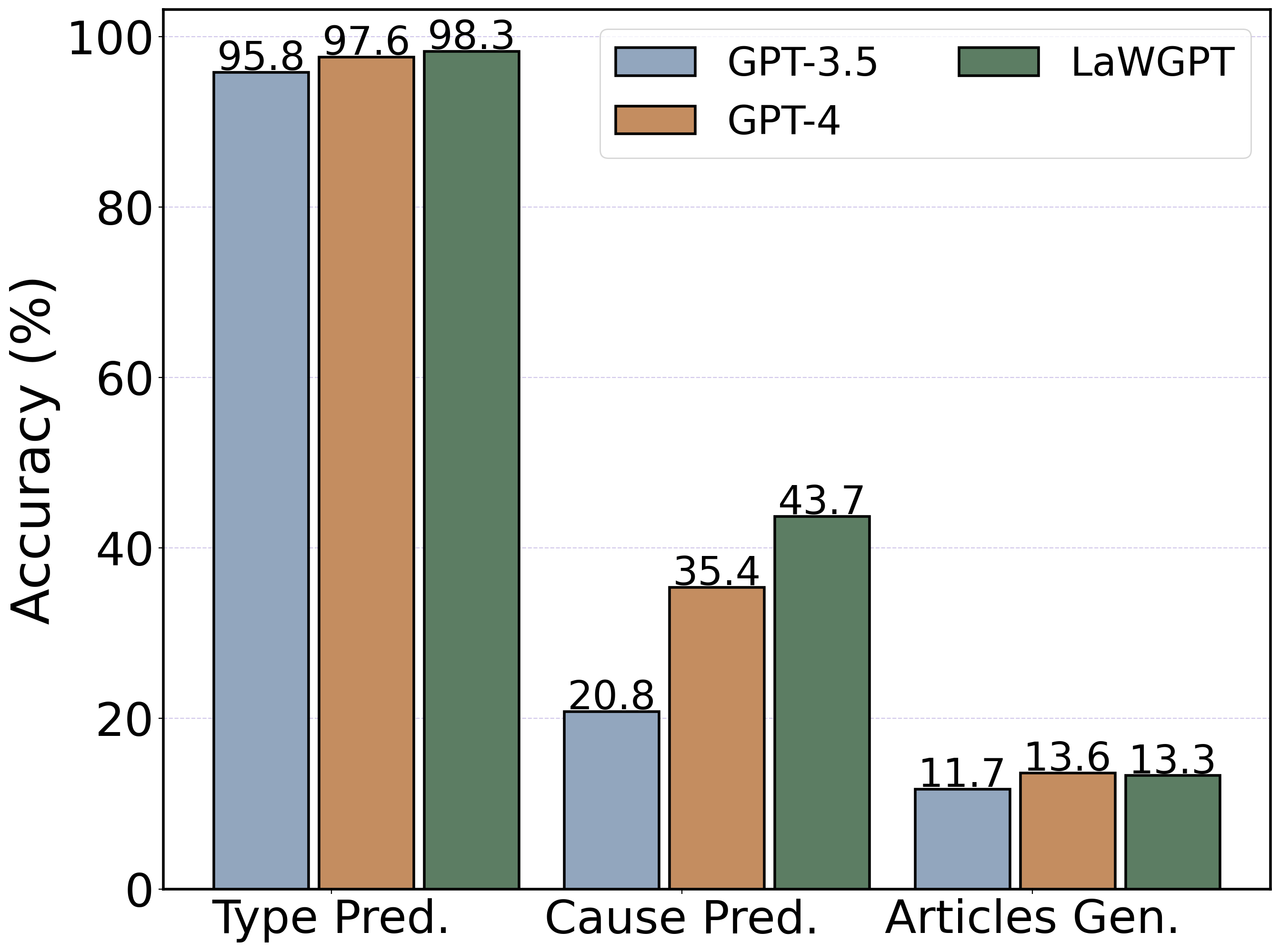}}  
	 	\caption{Legal knowledge evaluation of LLMs.}
   
	 	\label{fig:vanilla}
\end{figure}

\noindent \textbf{Legal Knowledge of LLMs} \quad As indicated in Figure \ref{fig:vanilla}, all three language models exhibit excellent performance on the simple task of predicting case types. However, their performance is less impressive on the challenging task of predicting case reasons, the GPT-4 model achieves only 35.4\% accuracy, while LaWGPT, which has undergone extensive pre-training with professional knowledge, achieves only 43.7\%. For the task of article generation, the performance of all models is poor, with LaWGPT sometimes producing garbled output, resulting in even worse performance.



\noindent \textbf{Multi-agent Court Simulation} \quad 
The results of the ablation experiments, as shown in Table \ref{tab:abla} in Appendix, demonstrate that our designed court debate simulation module effectively enhances the accuracy of judicial decisions. We further investigate the specific impact of this module on the prison term and fines in criminal case judgements. As depicted in Figure \ref{fig:change}, it is evident that the absolute difference in prison term and fines significantly diminishes following the simulation of court debates.

\noindent \textbf{Difficulty of Distinct Types of Cases} \quad Table \ref{tab:case_difficult} presents the results of our framework in generating legal articles across different types of cases in the first instance setting. The agent system produces more reliable legal articles in criminal cases, while its use and understanding of relevant legal statutes in civil and administrative cases are notably weaker. This observation may be attributed to the fact that the
civil and administrative cases involve more complex issues, with multiple vested interests, such as contract disputes, family matters, or government decisions, requiring a deeper understanding of legal and social knowledge.

\begin{table}
\centering
\resizebox{0.95\linewidth}{!}{
\begin{tabular}{cccc}
\toprule
\textbf{Case type} & \textbf{Precision} & \textbf{Recall} & \textbf{F1 Score} \\
\midrule 
\textbf{All} & 0.219 & 0.189 & 0.203 \\
\textbf{Criminal} & \textbf{0.489} & \textbf{0.264} & \textbf{0.343}\\
\textbf{Civil} & 0.073& 0.063& 0.067\\
\textbf{Administrative} & 0.126& 0.250& 0.167\\
\bottomrule
\end{tabular}}
\caption{Legal articles evaluation of AgentsCourt.}
\label{tab:case_difficult}
\end{table}

\noindent \textbf{Legal knowledge base} \quad With the support of an external knowledge base, the performance of agent system in judicial reasoning improved significantly, with an increase of up to 6.2\%. The achievements are also attributed to our designed automatic retrieval module. As shown in Table \ref{tab:retrieve} in Appendix \ref{sec:retrieve}, through the rough retrieval, the most similar cases only have a 62\% consistency in the cause of action with the current cases. However, after the documents re-ranking, the consistency of the cause of action between retrieved cases and the current cases increased to 85\%. This improvement proves the effectiveness of our retrieval module.

\section{Conclusion}

We propose a novel multi-agent framework AgentsCourt, which can sequentially simulate court debate, retrieve precedents, analyze cases, provide legal articles, and deliver clear judgment. Furthermore, we introduce SimuCourt, a judicial benchmark to evaluate the judicial analysis and decision-making power of agents. Then, we perform experiments to analyze different modules. The new judicial paradigm we presented effectively simulates the judicial decision-making with multi-agent, which significantly enhances judicial efficiency. 

\section{Limitation}


In this paper, we introduce a novel judicial benchmark SimuCourt. After thorough analysis, our work still presents the following limitations: 

\begin{itemize}
\item Our data only includes Chinese documents from "China Judgments Online." Despite our framewok AgentsCourt not being specifically designed for the civil law system, testing the agent system with real data from different legal systems is important.
\item The judgement documents cover the three most common types of cases: criminal, civil, and administrative. Including a broader range of case types in the future would evaluate the judicial analysis and decision-making power of agents more comprehensively.
\item Although our database contains a large number of precedents and legal resources, experimental results have shown that overall performance of agent systems is still unsatisfactory. 
\end{itemize}

We look forward to further exploring the potential of the legal knowledge base in future studies.


\bibliography{custom}

\appendix

\section{Retrieval Module}
\label{sec:retrieve}
As shown in Table \ref{tab:retrieve}, through the rough retrieval and documents re-ranking, the consistency of the cause of action between retrieved cases and the current cases increased to 85\%.

\section{Example of Court Transcript}
\label{sec:Transcript}
We present an example of court transcript simulated by multi-agent debate in Table \ref{tab:transcript}.

\section{Data Demonstration}
\label{sec:list}
\begin{figure*}[htbp]
	 	\centering  	\includegraphics[width=0.8\textwidth]{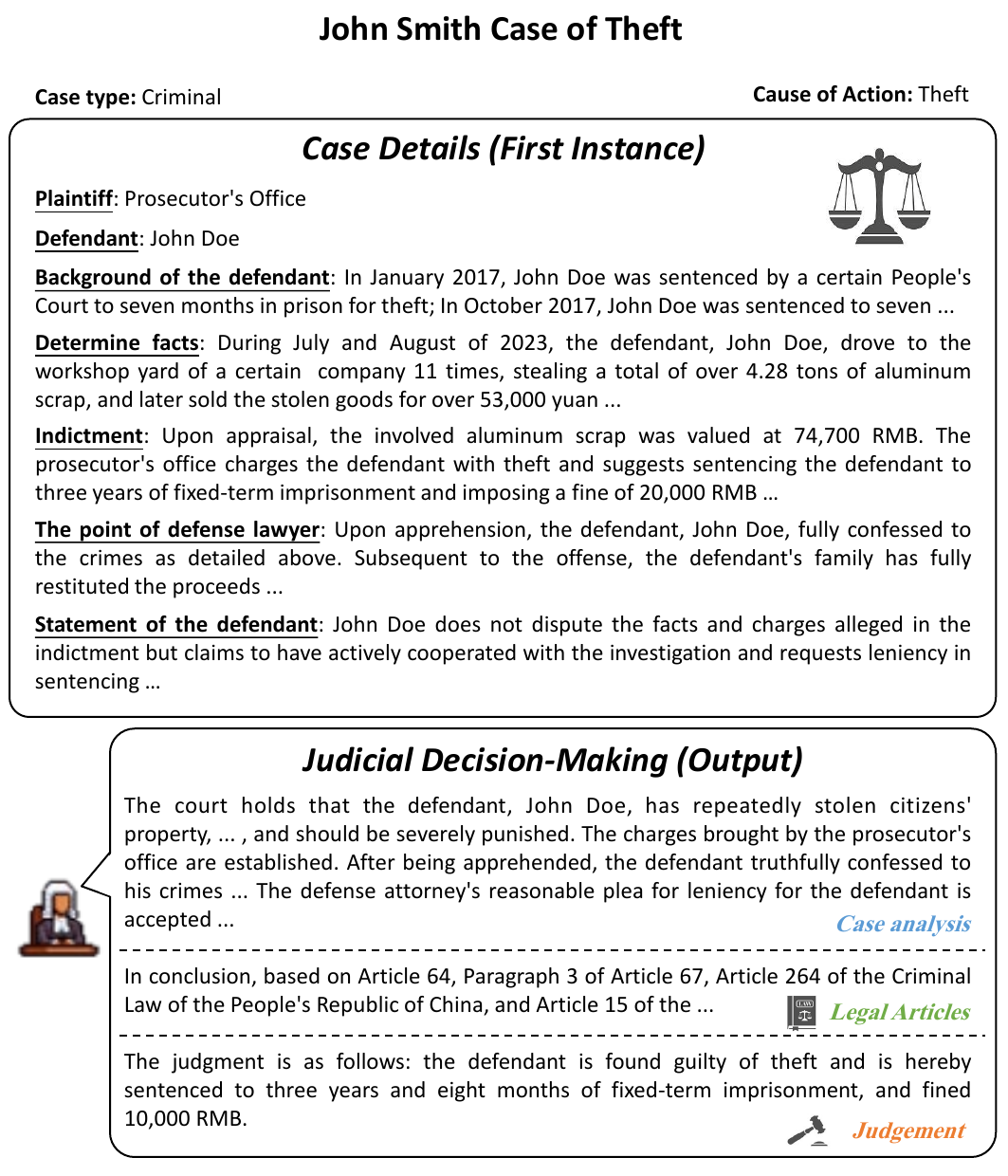}    
	 	\caption{An example case of first-instance stage (translated from Chinese).}
	 	\label{data_example1}
\end{figure*}

\begin{figure*}[htbp]
	 	\centering  	\includegraphics[width=0.8\textwidth]{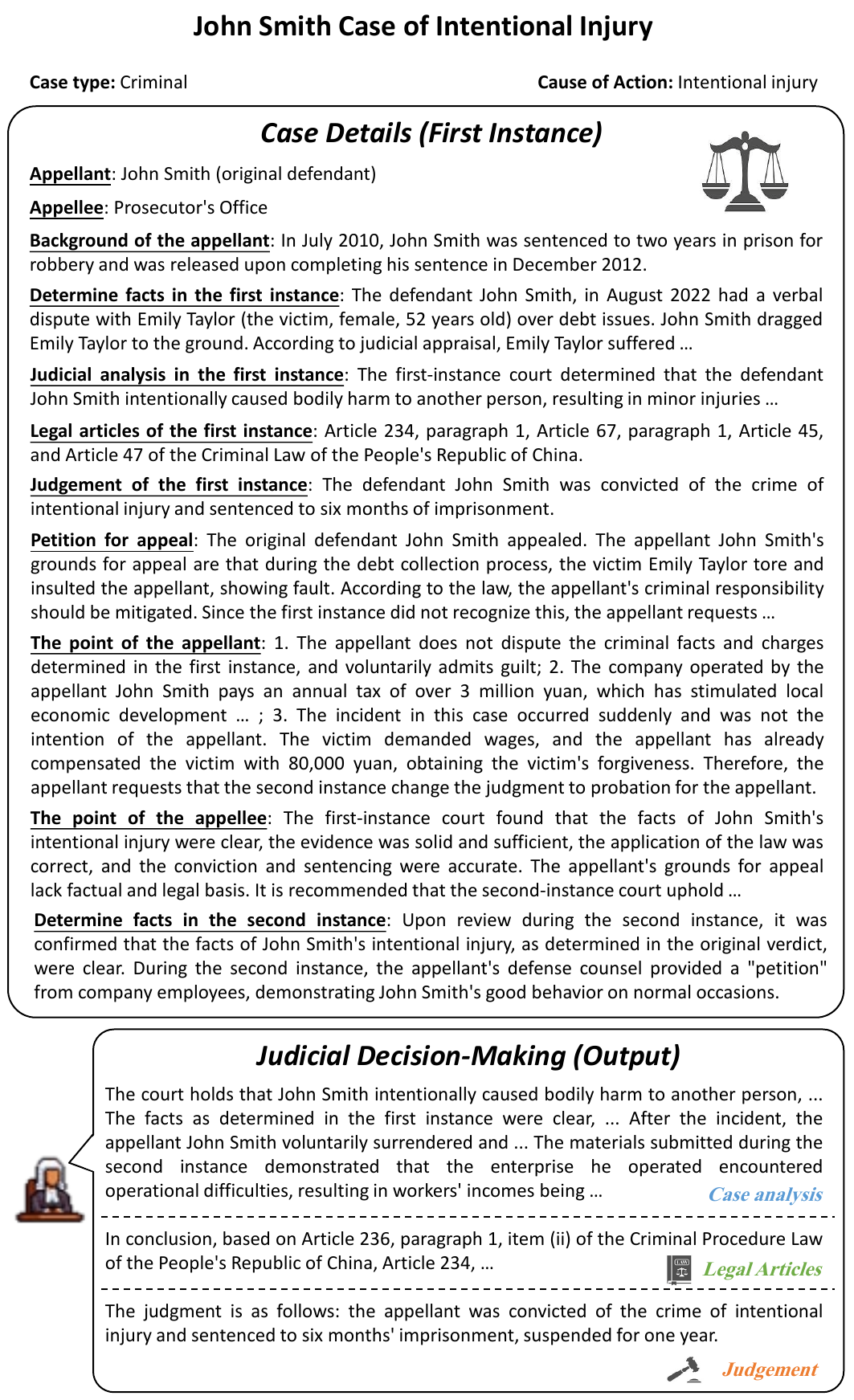}    
	 	\caption{An example case of second-instance stage (translated from Chinese).}
	 	\label{data_example2}
\end{figure*}

The detailed list is presented in Table \ref{tab:list}. Furthermore, we show examples of the first-instance stage in Figure \ref{data_example1} and second-instance stage in Figure \ref{data_example2}, respectively.

\section{Data Analysis}
\label{data_appendix}

\subsection{Data Description}

\begin{figure*}[htbp]
	 	\centering  	\includegraphics[width=1\textwidth]{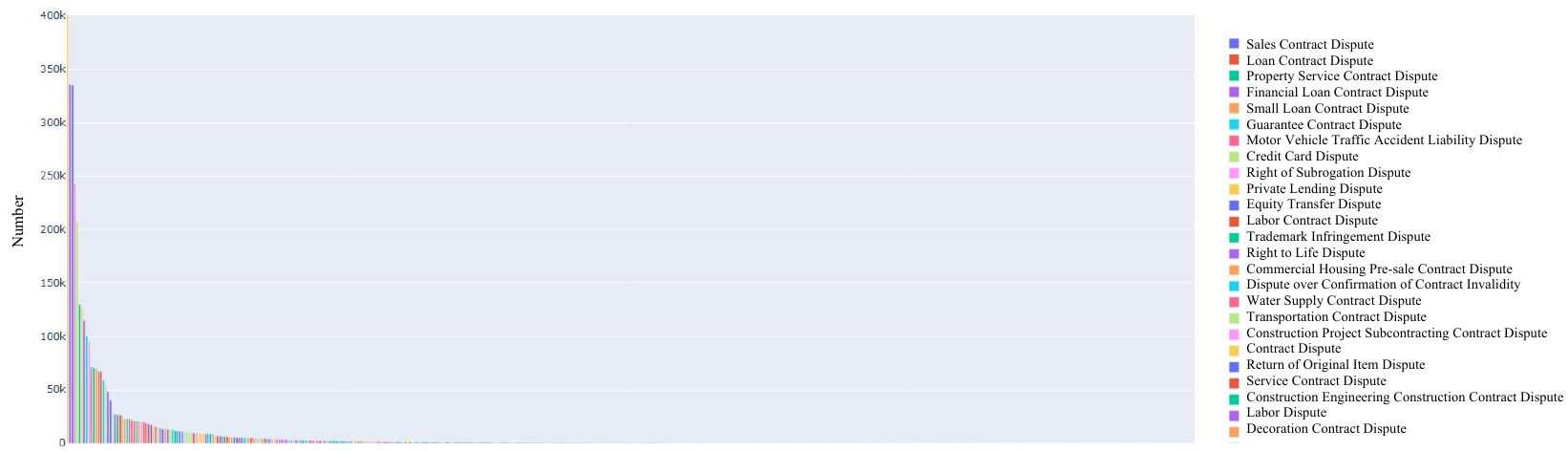}    
	 	\caption{Cause of action of civil cases  statistics in 2022}
	 	\label{data_analysis}
\end{figure*}

Our choice of cases is driven by three reasons: (1) \textbf{Diversity of causes of action}. Based on our statistical analysis of data from the China Judgements Online over the past few years, we observed a significant long-tail distribution in various types of cases. For example, as shown in Figure \ref{data_analysis}, in the total civil cases of 2022, the top 15 causes of action accounted for 66\% of the total number of cases. To reflect a broader spectrum of legal practice, we focus on maintaining diversity in the types of causes of action; (2) \textbf{Clarity of case analysis and facts}. We have meticulously selected judgement documents that provide detailed case analysis and clear determine facts for annotation. This aim is to enhance the quality and accuracy of data annotation while aiding agents in better understanding the judicial reasoning and legal articles; 
(3) \textbf{Uniqueness and accuracy of judgements}. We prioritize cases that are not overturned in appellate review. This ensures the consistency of our evaluation, as these cases have already undergone a rigorous litigation process and the judgements are fair. 

\subsection{Data Quality Inspection}
\label{sec:inspection}

We first process the privacy information of all documents. Specifically, We have meticulously anonymized sensitive information in the judgement documents. Then, After completing data annotation and handling private information, we manually inspect the data quality from various aspects. 

\noindent \textbf{Privacy Information Processing}: We have meticulously anonymized sensitive information in the judgement documents. In addition to replacing personal names, place names, and institution names with generic terms, we also anonymize other details that could potentially disclose personal privacy, such as ID numbers, phone numbers, and addresses, to ensure the safety of personal privacy.

\noindent \textbf{Manual Inspection}: After completing data annotation and handling private information, we manually inspect the quality of SimuCourt: (1) \textit{Case Meeting Standards}. The selected samples need to include clear case analysis and facts and have not been overturned in the appellate stage. (2) \textit{Accurate Information Annotation}. Annotation should ensure the accurate and error-free extraction of key information from the original legal documents, including case analysis, legal articles, and judgement. (3) \textit{Privacy Information Security}.In order to safeguard individual privacy and security, it is crucial to ensure that each data entry does not contain any content that could potentially disclose sensitive information about the parties involved. We employ three graduate students to manually review all 420 annotated cases. By carefully scrutinizing, our dataset exhibits a high level of quality. Specific quality metrics and analysis results are shown in Table \ref{tab:check}.

\begin{table}
\centering
\resizebox{\linewidth}{!}{
\begin{tabular}{cccc}
\toprule
\textbf{Precedents} & \textbf{Rough retrieval} & \textbf{ + Re-ranking} \\
\midrule 
Top1 & 62\% & 85\% \\
Top2 & 60\% & 82\% \\
Top3 & 61\% & 80\% \\
\bottomrule
\end{tabular}}
\caption{Cause of action matching}
\label{tab:retrieve}
\end{table}

\begin{table}
\centering
\resizebox{\linewidth}{!}{
\begin{tabular}{l|c}
\toprule
\textbf{Criteria} & \textbf{Pass Rate} \\
\midrule
Case Meeting Standards & 98.6\%\\
Accurate Information Extraction & 95.8\%\\
Privacy Information Security & 100\% \\ 
Average & 98.1\% \\
\bottomrule
\end{tabular}}
\caption{Data quality analysis.}
\label{tab:check}
\end{table}

\begin{figure}[!h]
	 	\centering{ 
	 	\includegraphics[width=0.49\textwidth]{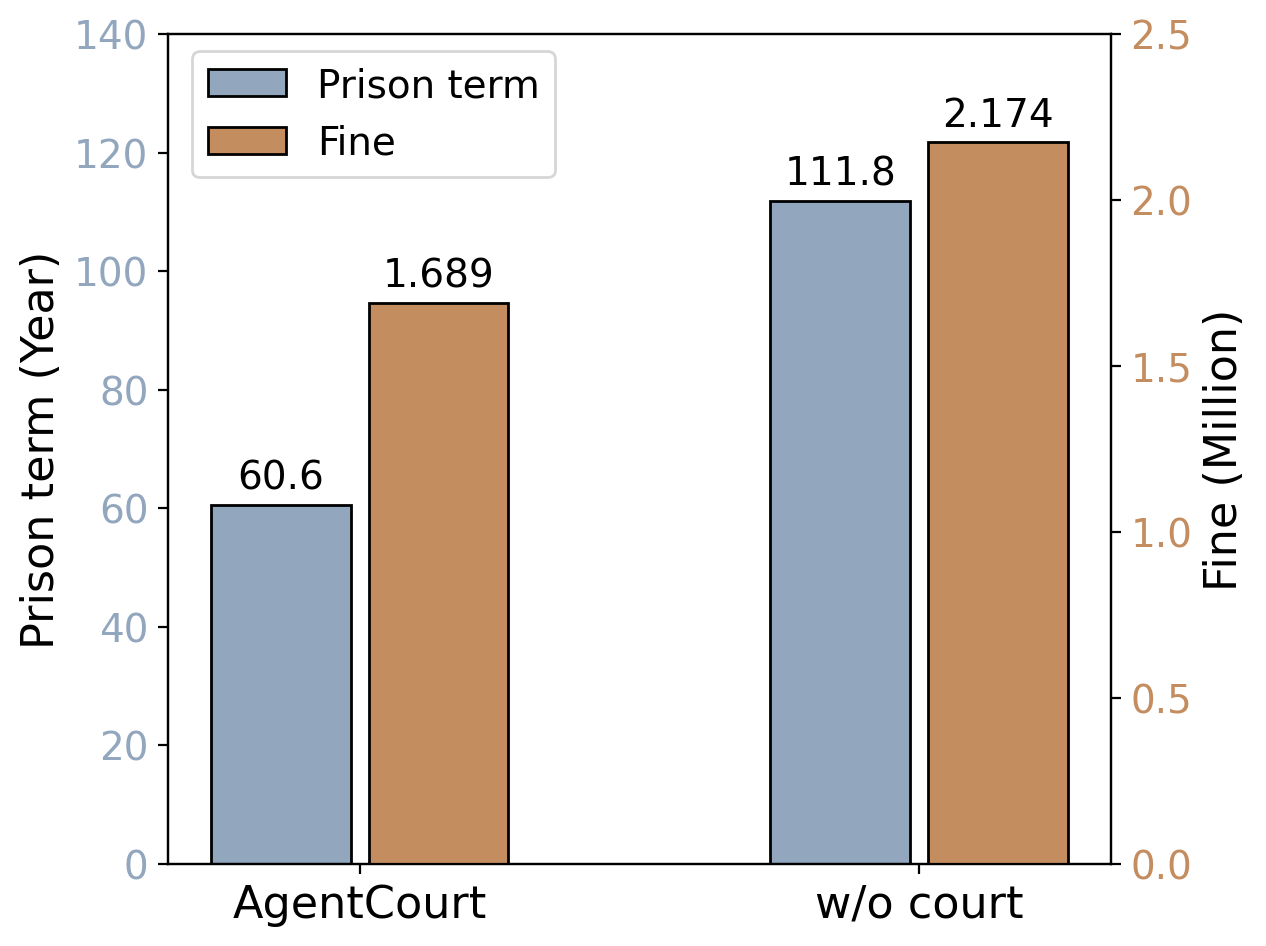}}  
	 	\caption{The absolute difference change.}
	 	\label{fig:change}
\end{figure}

\begin{table*}
\centering
\resizebox{\linewidth}{!}{
\begin{tabular}{ll}
\hline
\textbf{First instance} & \textbf{Second instance}\\

\hline
Case type & Case type \\
Cause of Action &  Cause of Action \\
Plaintiff & Appellant \\
  Defendant & Appellee \\
 Background information of the defendant	& Background information of the appellant\\
 Indictment	& Petition for appeal \\
The point of defense lawyer & The point of the appellant \\
The point of the defendant & The point of the appellee \\
Determine facts & Determine facts in the first instance \\
Case analysis  &  Judicial analysis in the first instance \\
Legal Articles  & Legal articles of the first instance \\
Judgement &   Judgement of the first instance \\
   &  Determine facts in the second instance \\
 &  Case analysis \\
  &  Legal Articles \\
 &  Judgement\\
\hline
\end{tabular}}
\caption{\label{tab:list}
Information list of different trial stages.
}
\end{table*}

\begin{table*}
\centering
\begin{tabular}{p{2.5cm}p{2.5cm}p{9.6cm}}
\toprule
\textbf{Cause of action} & \multicolumn{1}{c}{\textbf{Item}} & \multicolumn{1}{c}{\textbf{Content}}\\

\midrule
\multicolumn{1}{c}{\multirow{3}{*}{\textbf{Theft}}} & Case analysis & The court holds that the accused, John Doe, has repeatedly stolen citizens' property, constituting theft, and should be severely punished. The charges brought by the prosecutor's office are established. After being apprehended, the accused truthfully confessed to his crimes, voluntarily pleaded guilty, and returned part of the stolen goods, thus is eligible for a lighter punishment according to law. The defense attorney's reasonable plea for leniency for the accused is accepted.\\
\cmidrule(lr){2-3}
 & Legal articles & \textit{Article 64 of the Criminal Law of the People's Republic of China};
\textit{Paragraph 3 of Article 67 of the Criminal Law of the People's Republic of China};
\textit{Article 264 of the Criminal Law of the People's Republic of China};
\textit{Article 15 of the Criminal Procedure Law of the People's Republic of China}.\\
\cmidrule(lr){2-3}
 & Judgement & \textit{Charge}: The defendant is convicted of theft; \textit{Prison term}: Sentenced to three years and eight months in prison; \textit{Fine}: Fined ten thousand yuan.\\

\bottomrule
\end{tabular}
\caption{\label{tab:criminal}
An example output of criminal case.
}
\end{table*}

\begin{table*}
\centering
\begin{tabular}{p{2.5cm}p{12cm}}
\toprule
\multicolumn{1}{c}{\textbf{Role}} & \multicolumn{1}{c}{\textbf{Record}}\\
\midrule

 Judge & (1)Determine facts: On October 10, 2022, the defendant John Smith noticed the victim, David Lee, wearing a pair of gold earrings near a hotel. He then followed David Lee into an alley, and taking advantage of David Lee's unawareness, he snatched the pair of gold earrings from behind. After successfully obtaining them, the defendant John Smith fled the scene... (2) Background of the defendant: John Smith was sentenced to six months' imprisonment by a certain People's Court in June 2009 for the crime of theft; John Smith was sentenced to two years' imprisonment by a certain People's Court in December 2010 for the crime of robbery... (3) Court Discipline...The plaintiff and the defendant may now make their statements. 
\\
\midrule
Plaintiff & The defendant, John Smith, committed the crime of robbery with the purpose of unlawfully obtaining property from others, involving a considerable amount. His actions violate Article 267 of the Criminal Law of the People's Republic of China. The criminal facts are clear, and the evidence is indeed substantial and sufficient. Therefore, he should be held criminally liable for the crime of robbery.\\
\midrule
Defendant & I admit that the robbery occurred on October 10, 2022, which was an act of impulse. After the incident, I have truthfully confessed my crime. I hope the judge can give me a lenient punishment.\\
\midrule
Plaintiff & The defendant has previously been sentenced to a term of imprisonment for intentionally committing a crime. If he commits another crime within five years of completing the sentence, he should be sentenced to a term of imprisonment exceeding the original sentence, which constitutes recidivism.\\
\midrule
Defendant & I deeply acknowledge that my actions have caused adverse effects on society and irreversible losses to the victims. However, I assure the court that I will actively compensate the victims for their losses. I sincerely repent and am willing to accept the court's judgment.\\
\bottomrule
\end{tabular}
\caption{\label{tab:transcript}
An example of court transcript simulated by multi-agent debate (translated from Chinese).
}
\end{table*}

\section{Details of Automatic Evaluation}
\label{sec:auto_eval}
\subsection{Legal Articles Evaluation}
We start by using pattern matching to parse the free text, followed by a hard match against specific legal provisions. For example, as shown in Table \ref{tab:comp_article}. Then, with TP (True Positives) = 2, FP (False Positives) = 1, FN (False Negatives) = 2, the corresponding \textbf{\textit{Precision}} = 2/3, and \textbf{\textit{Recall}} = 2/4.

\subsection{Judgement Evaluation for Civil and Administrative Cases}
We utilize GPT-4 to assess the judgment results generated by the model in civil and administrative cases. As shown in Table x, we present an evaluation example, which is also a prompt demonstration for GPT-4.

\begin{table*}
\centering
\begin{tabular}{p{7.25cm}p{7.25cm}}
\toprule
\multicolumn{1}{c}{\textbf{Reference legal articles}} & \multicolumn{1}{c}{\textbf{Generated legal articles}}\\
\midrule

 Article 67, Section 1 of the Criminal Law of the People's Republic of China & Article 67 of the Criminal Law of the People's Republic of China
\\
\midrule
Article 52 of Criminal Law of the People's Republic of China & Article 53 of the Criminal Law of the People's Republic of China \\
\midrule
Article 53 of the Criminal Law of the People's Republic of China & Article 52 of Criminal Law of the People's Republic of China \\
\midrule
 Article 15 of the Criminal Procedure Law of the People's Republic of China &  \\

\bottomrule
\end{tabular}
\caption{\label{tab:comp_article}
A comparison between  the reference legal articles and the legal articles generated by the model (translated from Chinese).
}
\end{table*}

\begin{table*}
\centering
\begin{tabular}{p{7.25cm}p{7.25cm}}
\toprule
\multicolumn{1}{c}{\textbf{Reference judgment}} & \multicolumn{1}{c}{\textbf{Generated judgment}}\\
\midrule

 \textbf{Result1}: The defendant, Company XYZ, shall return the plaintiff's rental deposit of 6000 yuan and the remaining rent of 17025 yuan within five days from the effective date of this judgment. & \textbf{Result1}: The defendant shall return the deposit of 6000 yuan to the plaintiff.
\\
\midrule
\textbf{Result2}: The defendant, Company XYZ, shall pay the plaintiff a penalty of 5950 yuan within five days from the effective date of this judgment. & \textbf{Result2}: The defendant shall return the remaining rent of 17025 yuan to the plaintiff. \\
\midrule
\textbf{Result3}: The plaintiff's other claims are dismissed &  \\
\midrule

\midrule
 \textbf{The analysis by GPT-4:} &  \\
\midrule
 \multicolumn{1}{c}{\textbf{Key Point Match Evaluation}} & The standard answer contains 4 key points: the return of the rental deposit of 6000 yuan, the return of the remaining rent of 17025 yuan, the payment of a penalty of 5950 yuan, and the dismissal of other claims. The generated answer's "Result 1" matches the first key point (the return of the rental deposit of 6000 yuan), and "Result 2" matches the second key point (the return of the remaining rent of 17025 yuan). Therefore, the score for key point match is 2. \\
\midrule
 \multicolumn{1}{c}{\textbf{Answer Accuracy Evaluation}} & The generated answer consists of 2 results. The "Result 1" of the generated answer accurately matches one key point without including incorrect information, thus scoring 1 point. Similarly, the "Result 2" of the generated answer accurately matches another key point without including incorrect information, also scoring 1 point. Therefore, the score for accuracy is 2. \\
\midrule
 \multicolumn{1}{c}{\textbf{Output Result}} & Standard Key Points Count: 4, Key Point Match Score: 2, Generated Answer Results Count: 2, Answer Accuracy Score: 2 \\
\midrule
 \multicolumn{1}{c}{\textbf{Precision}} & Answer Accuracy Score / Generated Answer Results Count = 2/2 \\
\midrule
 \multicolumn{1}{c}{\textbf{Recall}}  & Key Point Match Score / Standard Key Points Count = 2/4 \\

\bottomrule
\end{tabular}
\caption{\label{tab:GPT_eval}
A prompt demonstration for GPT-4 evaluator (translated from Chinese).
}
\end{table*}

\end{document}